\pdfoutput=1

\PassOptionsToPackage{numbers, compress}{natbib}

\documentclass[11pt]{article}

\usepackage[preprint]{neurips_data_2024}

\usepackage{times}
\usepackage{latexsym}
\usepackage{hyperref}
\hypersetup{
    colorlinks=true,
    }
    
\usepackage[utf8]{inputenc} 
\usepackage[T1]{fontenc}    
\usepackage{booktabs}       
\usepackage{amsfonts}       
\usepackage{nicefrac}       
\usepackage{microtype}      
\usepackage{xcolor}         

\usepackage{todonotes}
\usepackage{pgfplots}
\usepackage{multirow}
\usepackage{markdown}
    
\usepackage{graphicx}
\usepackage{subcaption}

\usepackage[T1]{fontenc}

\usepackage[utf8]{inputenc}

\usepackage{microtype}
\usepackage{enumitem}

\usepackage{inconsolata}

\newcommand{\datalink}{\href{https://github.com/google-research/google-research/tree/master/youtube_sl_25}{this link}}

\title{YouTube-SL-25: A Large-Scale, Open-Domain Multilingual Sign Language Parallel Corpus}

\author{Garrett Tanzer\thanks{Correspondence to \texttt{gtanzer@google.com}.} \\
  Google \\
  \And Biao Zhang \\
  Google DeepMind}

\begin{document}
\maketitle
\begin{abstract}
Even for better-studied sign languages like American Sign Language (ASL), data is the bottleneck for machine learning research. The situation is worse yet for the many other sign languages used by Deaf/Hard of Hearing communities around the world. In this paper, we present YouTube-SL-25, a large-scale, open-domain multilingual corpus of sign language videos with seemingly well-aligned captions drawn from YouTube. With $>$3000 hours of videos across $>$25 sign languages, YouTube-SL-25 is a) $>$3x the size of YouTube-ASL, b) the largest parallel sign language dataset to date, and c) the first or largest parallel dataset for many of its component languages. We provide baselines for sign-to-text tasks using a unified multilingual multitask model based on T5 and report scores on benchmarks across 4 sign languages. The results demonstrate that multilingual transfer benefits both higher- and lower-resource sign languages within YouTube-SL-25.\footnote{We release the YouTube-SL-25 video IDs at \datalink.}
\end{abstract}

\section{Introduction}

There are \href{https://www.un.org/en/observances/sign-languages-day}{>300 sign languages} used by Deaf/Hard of Hearing communities around the world. As minority languages, sign languages have relatively little data available and therefore---like low-resource spoken languages---are challenging to process with machine learning. The fact that sign languages are visuospatial languages represented as video with no commonly used written form creates extra challenges at every stage of development: they are more difficult to mine, filter, preprocess, and model. Datasets like YouTube-ASL~\citep{youtubeasl} for American Sign Language (ASL) and BOBSL~\citep{bobsl} for British Sign Language (BSL) make a focused effort to advance the status quo for a single sign language, but many of the world's sign languages are being left behind.

In this paper, we present YouTube-SL-25, a large-scale, open-domain corpus of multilingual sign language videos with seemingly well-aligned captions, primarily intended for translation from each sign language to its region's spoken language. By \textit{open-domain}, we are distinguishing from datasets like AfriSign~\citep{gueuwou2023afrisign} and JWSign~\citep{gueuwou2023jwsign}, which have made strides towards massive sign-multilinguality but feature a limited domain (e.g., Bible translations).

We mined these videos using a two-step process: First, like YouTube-ASL~\citep{youtubeasl}, we used automatic classifiers on text metadata to identify potentially relevant videos. And second---in contrast to YouTube-ASL, which paid 3 native ASL signers to filter out individual videos with wrong or misaligned captions over the course of several months, a process that isn't amenable to scaling across many sign languages---we used our own knowledge of sign languages and YouTube data to triage the videos over four days, ranking the priority of channels according to total hours of content and then auditing videos per channel, with particular attention to outliers by duration. This means that the annotations were performed with less expertise than YouTube-ASL, but in practice there are many signals that can be used to identify high-quality content even without complete understanding, as observed in work with written languages~\citep{representationWashing, kudugunta2023madlad400}.

The result is a dataset with 3207 hours of videos with seemingly well-aligned captions, featuring $>$3000 unique signers across $>$25 sign languages. This is $>$3x the size of YouTube-ASL (984 hours), which is a subset of YouTube-SL-25, and larger than the prior largest parallel sign language dataset, JWSign (2530 hours), which is closed-domain. YouTube-SL-25 is the first or largest parallel dataset for many of its component sign languages.

Following~\citet{representationWashing}'s call to avoid ``representation washing''---i.e., overstating the number of low-resource languages supported when their dataset size is marginal---we name YouTube-SL-25 after the $>$25 sign languages that have at least 15 hours of representation in it. This is roughly the minimum size of datasets that have previously been released for individual sign languages, but it is still considered extremely low resource. When including the long tail, YouTube-SL-25 includes data in at least 55 sign languages.

We provide baselines for YouTube-SL-25 on sign language understanding tasks, extending the multitask mixture from FLEURS-ASL's baselines~\citep{fleursasl} to support multiple source/target languages and the sign language identification task (which is extremely understudied~\citep{gebre2013,gebre-etal-2014-unsupervised,SultanMakramKayedAli+2022+191+210}). Our results on benchmarks across 4 sign languages show that both higher- and lower-resource sign languages within YouTube-SL-25 benefit from multilingual transfer.

We publicly release the YouTube-SL-25 video IDs at {\datalink}. We hope that YouTube-SL-25 can be further refined by language experts in the community and will be useful for a variety of purposes, such as general sign language pretraining and medium-quality finetuning for downstream tasks like translation, caption alignment, \& sign language identification. In particular, we think that it is necessary to develop robust filtering and preprocessing tools for sign languages in order to realize the next order of magnitude increase in dataset size.

\begin{figure}
    \centering
    \includegraphics[scale=0.6]{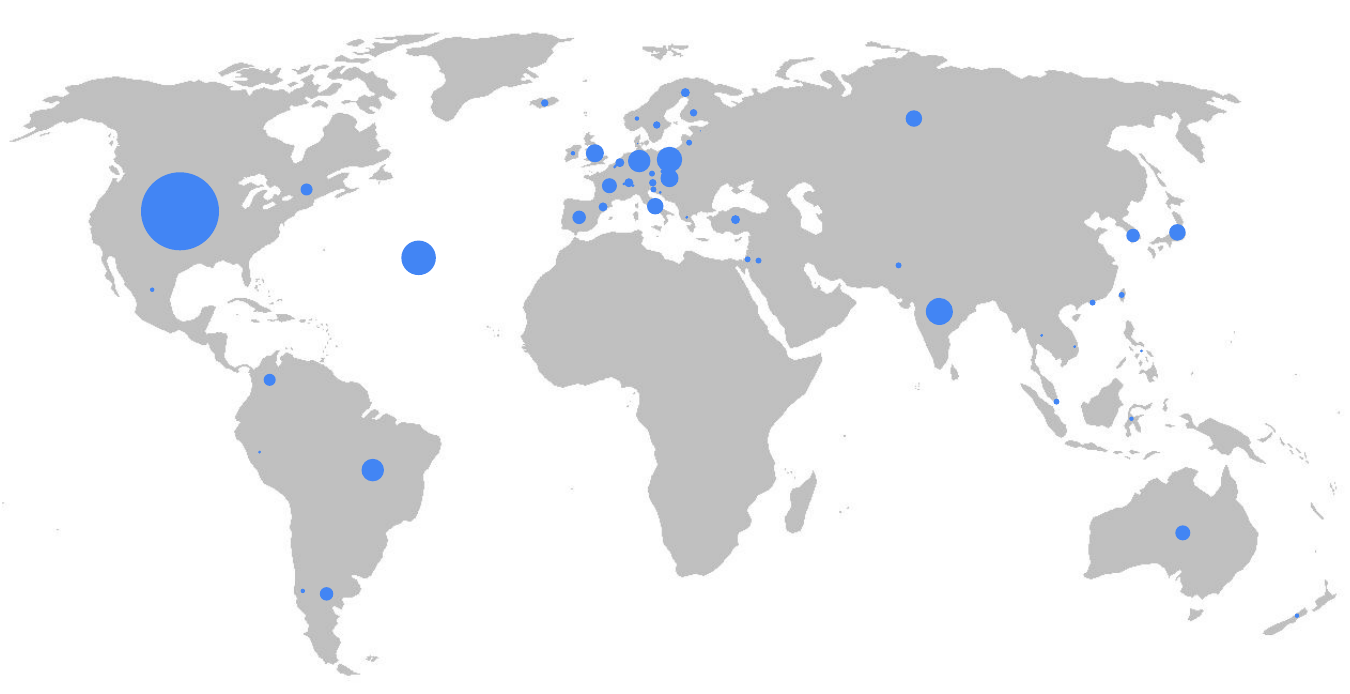} 
    \caption{\textbf{\href{https://commons.wikimedia.org/wiki/File:BlankMap-World-noborders.png}{World map} showing the amount of content in YouTube-SL-25 for each sign language;} the area of each circle is proportional to the number of hours. The circle in the middle of the Atlantic Ocean represents International Sign. Observe that the dataset is especially lacking in representation for Central \& South America, Africa, West \& Central Asia, and China \& Southeast Asia.}
    \label{fig:worldmap}
\end{figure}

\section{The YouTube-SL-25 Corpus}
\label{sec:corpus}

YouTube-SL-25 is a massively multilingual corpus of sign language videos with seemingly well-aligned captions drawn from YouTube, intended primarily for pretraining sign-to-text translation models. YouTube-SL-25 is a superset of YouTube-ASL~\citep{youtubeasl}.

YouTube-ASL used a two-step pipeline to construct its corpus: ``first, retrieval using automatic content-based annotations, and second, filtering by skilled human annotators at a per-video level.'' We adopt the first step with minimal modifications, but use a lower standard for the second step: triaging/auditing by a single non-native signer to identify seemingly high-quality channels for inclusion (even without full understanding of the content in every sign language).

\subsection{Automatically Retrieving Candidate Videos}
\label{subsec:retrieve}

We retrieved candidate videos in a similar way to YouTube-ASL~\citep{youtubeasl}, by fetching videos tagged with Knowledge Graph entities related to sign language generally or any individual sign language with its own tag as of July 2023.\footnote{We refreshed the dataset in May 2024 to include newer videos from accepted channels, with some light manual review.} As with YouTube-ASL, the recall of these tags is limited in that they are not aware of sign language in the video content itself (i.e., they do not use sign language detection and sign language identification classifiers), so they may miss videos or channels that use sign language but do not explicitly mention it. The pool of videos that we fetched from is also restricted in some generic ways, such as that the videos must be public and listed.

We applied the same filtering steps as YouTube-ASL: select only videos with manually uploaded captions, and remove videos with duration $<$10 seconds or $>$5 hours, width $<$480 pixels or height $360$ pixels, and frame rate $<$15fps or $>$60fps. We added one additional step: remove videos where captions cover $<$40\% of the duration.

Unlike YouTube-ASL, we did not exclude videos of conversations with more than one signer in frame at a time. While our pose-based baselines only support one person's input, we feel that there is no reason to exclude these videos from the corpus itself. We are not aware of any prior works that study sign language translation from videos with multiple signers; some datasets like the Public DGS Corpus~\citep{hanke-etal-2020-extending} consist of conversational data, but they are recorded with one camera per signer. Being able to handle this kind of data is important both in the near term to make the most of the limited resources available and in the long term to support multi-signer applications.

The result of this process was a list of 81,623 candidate videos that might contain signed content with high-quality captions.

\subsection{Triaging Candidate Videos with Coarse Manual Annotations}
\label{subsec:filter}

YouTube-ASL used 3 native ASL signers as annotators to identify which of the individual candidate videos contained ASL and had high-quality, well-aligned English captions, labelling videos over the course of several months and several rounds of iteration on the annotation tool/labelling standards. This kind of annotation does not scale well to massively multilingual data because it is difficult and expensive to onboard native (or even just proficient) signers to annotate videos in tens of (extremely low resource) sign languages. Instead, we follow~\citet{representationWashing}'s observation that many data filtering tasks can be done even without full understanding of the content. Therefore, the first author (a non-native hearing signer with experience in several sign languages) served as the annotator, triaging the corpus over the course of a four-day weekend using a combination of per-channel annotations and targeted audits. The triage proceeded as follows:

First, we automatically included the YouTube-ASL videos in our corpus and removed these from the manual review. Then we grouped videos by channel and sorted by total duration in descending order, with random ordering for videos within each channel. We used heuristic classifiers based on public text metadata for each video and channel (e.g., title, description, caption language) to provisionally label each video with a sign language. For each channel, the author sampled several random videos and judged whether they matched the acceptance criteria (described below), then corrected the sign language label if relevant. If all the sampled videos were good, the channel was tentatively accepted; if the videos were bad on balance, it was rejected; if it was mixed but promising, the author investigated further and was able to exclude individual bad videos. The author triaged the first 1000 channels of 18000 (down to about 1 hour of content per channel), then continued into the 3000-6000 range, skipping videos predicted to be in ASL in order to prioritize the long tail of sign languages.\footnote{An observation: the acceptance rates for videos in different sign languages were dramatically different. For example, the vast majority of content from Central \& South America was live-interpreted events with small interpreters and poor caption alignment. We expect that high-quality produced content is more common in high-income countries, so it will be important for future work to leverage lower quality data in order to be more globally inclusive.} Finally, we sorted the videos for each channel by duration in descending order and the first author made a second pass, checking outlier videos with unusually long durations for their channels\footnote{Sometimes channels with high-quality produced content with relatively consistent duration $\sim$10 minutes would have long live-interpreted events with different translation and caption quality.} to reduce heavy-hitter errors.

Now, a description of our annotation standard for individual videos and some shallow signals to recognize these features even without full understanding:

\begin{itemize}[leftmargin=6mm]
    \item The video should feature sign language. While there is such a thing as \href{https://www.nbcnews.com/news/world/fake-sign-language-interpreter-nelson-mandela-memorial-provokes-anger-flna2d11723934}{fake sign language interpreters}, they do not typically create YouTube channels full of themselves gesturing nonsensically, so there is not much adversarial content to sift through. The closest thing to this is inexperienced sign language learners, especially producing loose translations of music.\footnote{We accept learners as long as they seem relatively competent. Models should be able to understand less proficient signers, especially given that many deaf people are not exposed to sign language in childhood and reach varying levels of proficiency~\citep{signExposurePct}. Less proficient signing can be easily distinguished from native signing, so it should not harm the quality of sign language understanding (especially if the data is only used for pretraining), but more care is required for generation.} An annotator proficient in one sign language can generally distinguish these cases (and others like interpretive dance/gesture) even for other sign languages.
    \item The signer(s) should be the primary content. If a signer is interpreting spoken content, the video of the speaker (if present) should be significantly smaller than the signer. If the person is switching off between signing and speaking, or a speaker is the primary content for a significant fraction of the video, it should be rejected.
    \item The video should have relatively well-aligned captions. There are a few signals you can use for this: if the video has no speech track, the captions are virtually always well-aligned;\footnote{These are typically channels with natively Deaf-produced content.} you can validate this by seeing whether the changes in captions occur at boundaries in sign language phrases (which are marked pretty universally with pauses and nonmanual signals like head tilt or eyebrow movement). If there is a speech track, you can listen to whether the caption timing matches the speech track. We accepted videos where the captions were aligned to the speech track as long as they were not off by more than a couple of seconds; usually this corresponds to an after-the-fact voiceover translation, not live interpretation.
    \item Label which sign language the video is in. It was almost always possible to classify the sign language with high confidence based on the video title/description and caption language, or at worst the channel's description (including its stated country). The failures in automatic classification mostly came from International Sign, often specified as just ``IS'', which was not automatically tagged. In the rare cases where we could not determine the language with high confidence, we marked the sign language as unknown.
\end{itemize}

In order to sanity check the end-to-end quality of this triage process, the first author ran two additional audits on the final dataset (excluding the videos from YouTube-ASL, since we were already confident in their quality). The author rated two samples of 100 videos, one sampled uniformly per video id, and the other with videos sampled with probability proportional to their video duration. The author loaded each video and scrubbed through the frame previews across time to identify cuts where content and captioning habits may change.

In the sample with random videos by ID, we found: 1 video that strictly should not have been included in the dataset (a brief video with captioned narration), 1 borderline video with a waist-up interpreter next to a video of a speaker's face only, and 3 videos with sign-along versions of songs that used appropriate vocabulary but didn't really provide a complete translation.

In the sample proportional to video duration, we found: 0.76\% of content that was speech rather than signing (from a channel with otherwise entirely signed content), and 0.96\% that consisted of interstitials mixed within otherwise good content, where the interstitials had signing but no captions.

We consider these error rates acceptable.

\begin{figure}[t]
    \centering
    \begin{subfigure}{0.49\textwidth}
    \centering
    \includegraphics[scale=0.38]{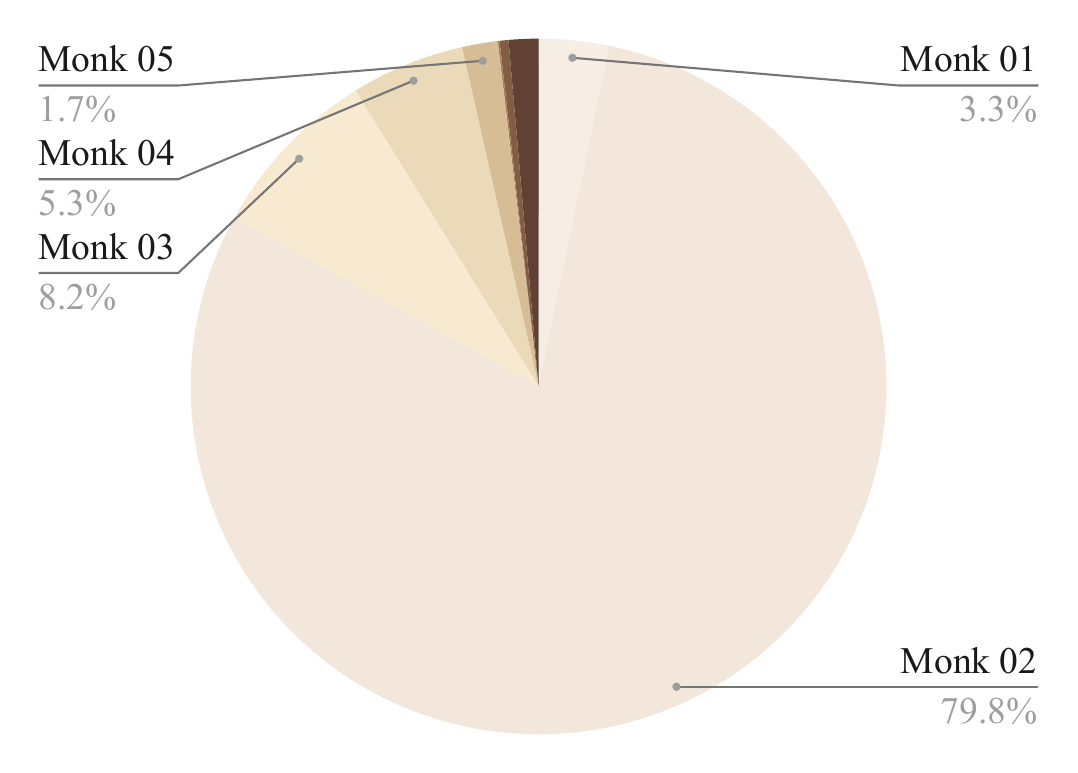} 
    \caption{\textbf{Monk skin tone ratings}~\citep{Monk_2019}. Dark skin is globally underrepresented; some of this is due to the regional bias towards the Global North.}
    \end{subfigure}\hfill%
    \begin{subfigure}{0.49\textwidth}
    \centering
    \includegraphics[scale=0.38]{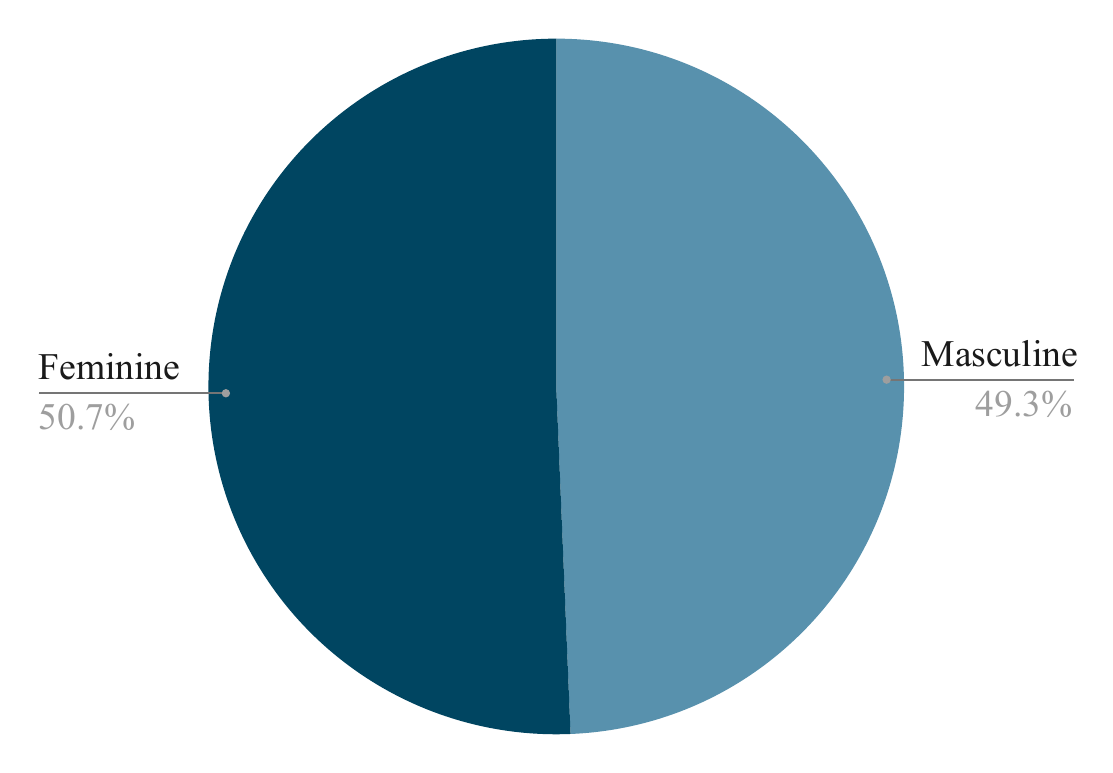}
    \caption{\textbf{Perceived gender presentation} (nonbinary presentation not supported by the classifier). Masculine and feminine presentation are at parity.}
    \end{subfigure}
    \caption{\textbf{Demographic representation of YouTube-SL-25 content (proportion of hours)}, predicted with proprietary classifiers. These predictions should only be interpreted in aggregate.}
    \label{fig:fairness}
\end{figure}

\subsection{Corpus Statistics \& Comparison to Prior Datasets}
\label{subsec:stats}

The final human-triaged YouTube-SL-25 corpus consists of 39197 videos totaling 3207 hours of content (covered by 2980 hours of captions = 2.16M captions = 104M characters)\footnote{We used classifiers to evaluate whether the captions were toxic, hateful, violent, and sexual with confidence >0.8, which flagged <1\% of captions. We manually audited the captions flagged with highest confidence and none merited exclusion from the dataset: the ``toxic'' and ``hateful'' content were false positives, the ``violent'' content was news about violent events, and the ``sexual'' content was news or educational content about sex.} across $>$25 sign languages.mom Like in YouTube-ASL, we approximately lower bound the number of unique signers using the number of unique channels, 3072, and expect that this is quite an underestimate as channels (or even individual videos) featuring multiple signers are common. This is only somewhat larger than YouTube-ASL's 2519 channels, which is the main downside of our triage-based annotation approach: it captures the bulk of the hours but not the long tail of unique channels.

As discussed in~\citet{youtubeasl}'s related work section, to the best of our knowledge YouTube-ASL and now YouTube-SL-25 are the only sign language translation datasets that aggregate content from a wide variety of automatically mined channels, as opposed to others which either pay participants to create new data~\citep{how2sign,voskou2023new,ko2019neural,hanke-etal-2020-extending} or curate content from a small number of content creators~\citep{openasl,bobsl,shen2023auslandaily,ouakrim:hal-04494094,gueuwou2023jwsign,joshi2023isltranslate,wmt_slt_23}. This has implications for diversity in topics/content and video production style.\footnote{Note that even though YouTube-SL-25 is open-domain overall, there may be discretization effects in the long tail of languages, where a language is represented by only a few channels and therefore in effect is closed-domain.}


See Figure~\ref{fig:worldmap} for a depiction of the relative weight of each language on a world map and Table~\ref{fig:dataset-hours} for a breakdown of hours by language, comparing to the largest prior parallel dataset for each language (both open- and closed-domain). Note that prior works have studied sign languages with no representation in YouTube-SL-25, like AfriSign~\citep{gueuwou2023afrisign} \& JWSign~\citep{gueuwou2023jwsign} (primarily languages in Central \& South America and Africa), and CSL-Daily~\citep{zhou2021improving} (China); these languages aren't included in the table. The distribution of language data can be attributed to population size, economic development, and political factors. Other supplementary sources of data, such as Bible translations~\citep{gueuwou2023afrisign,gueuwou2023jwsign} and national interpreted broadcasts~\citep{bobsl,camgoz2021content4all}, could provide more balanced data distributions.

See Figure~\ref{fig:fairness} for an estimated demographic breakdown of YouTube-SL-25 by skin tone and perceived gender presentation, predicted by automated classifiers. Dark skin is globally underrepresented in the dataset, with only 1.9\% (60 hours) of the data at Monk 6-10. Some of this can be explained by overrepresentation of countries in the Global North (and the only two countries in the Global South with a significant fraction of the dataset, Brazil and India, have substantial variation in skin tone). Ensuring demographic fairness remains an important problem for future work.


\begin{table*}
    \centering
    \small
    \setlength{\tabcolsep}{4pt}
    \begin{tabular}{lcccc|lc|lc}
    \toprule
    \bf sign language & \bf iso 639 & \bf \#videos & \bf \#channels & \bf \#hours & \bf \begin{tabular}{@{}c@{}}largest prior \\ (open) \end{tabular} & \bf \#hours & \bf \begin{tabular}{@{}c@{}}largest  \\ prior \end{tabular} & \bf \#hours \\
    \midrule
    American & \tt ase & 16724 & 2523 & \bf 1394 & YouTube-ASL~\citep{youtubeasl} & 984 & \textquotedbl & \textquotedbl \\
    International & \tt ils & 1634 & 14 & \bf 285 & - & - & - & - \\
    Indian & \tt ins & 3023 & 8 & \bf 209 & ISLTransate~\citep{joshi2023isltranslate} & 55 & \textquotedbl & \textquotedbl\\
    Polish & \tt pso & 1698 & 34 & \bf 137 & - & - & JWSign & 61 \\
    German & \tt gsg & 1024 & 65 & \bf 108 & DGS Corpus~\citep{hanke-etal-2020-extending} & 50 & \textquotedbl & \textquotedbl \\
    Brazilian & \tt bzs & 846 & 30 & 101 & - & - & JWSign & \bf 211 \\
    British & \tt bfi & 1026 & 60 & 74 & BOBSL~\citep{bobsl} & \bf 1660 & \textquotedbl & \textquotedbl \\
    Hungarian & \tt hsh & 1687 & 9 & \bf 70 & - & - & JWSign & 16 \\
    Australian & \tt asf & 1098 & 25 & \bf 67 & Auslan-Daily~\citep{shen2023auslandaily} & 45 & \textquotedbl & \textquotedbl \\
    Italian & \tt ise & 929 & 21 & 63 & - & - & JWSign & \bf 115 \\
    Japanese & \tt jsl & 1075 & 21 & 62 & - & - & JWSign & \bf 67 \\
    Russian & \tt rsl & 715 & 9 & 60 & - & - & JWSign & \bf 111 \\
    French & \tt fsl & 900 & 23 & 49 & Mediapi-RGB~\citep{ouakrim:hal-04494094} & \bf 86 & \textquotedbl & \textquotedbl \\
    Korean & \tt kvk & 325 & 10 & 38 & - & - & JWSign & \bf 93 \\
    Colombian & \tt csn & 212 & 4 & 37 & - & - & JWSign & \bf 128 \\
    Spanish & \tt ssp & 701 & 13 & 36 & - & - & JWSign & \bf 85 \\
    Dutch & \tt dse & 450 & 14 & \bf 35 & - & - & JWSign & 1 \\
    Argentine & \tt aed & 160 & 9 & 34 & - & - & JWSign & \bf 107 \\
    Quebec & \tt fcs & 128 & 7 & \bf 26 & - & - & JWSign & 20 \\
    Catalan & \tt csc & 239 & 7 & \bf 26 & - & - & - & - \\
    Pakistani & \tt pks & 581 & 1 & \bf 25 & - & - & - & - \\
    Swedish & \tt swl & 410 & 23 & \bf 22 & - & - & JWSign & 14 \\
    Turkish & \tt tsm & 189 & 8 & 18 & E-TSL~\citep{ozturk2024etsl} & \bf 24 & \textquotedbl & \textquotedbl \\
    Swiss German & \tt sgg & 185 & 6 & 18 & SRF23~\citep{wmt_slt_23} & \bf 437 & \textquotedbl & \textquotedbl \\
    Israeli & \tt isr & 345 & 14 & \bf 17 & - & - & JWSign & 1 \\
    Fenno-Swedish & \tt fss & 174 & 8 & \bf 16 & - & - & - & - \\
    Finnish & \tt fse & 196 & 13 & 14 & Corpus FinSL~\citep{salonen-etal-2020-corpus} & \bf 15 & JWSign & 35 \\
    Austrian & \tt asq & 177 & 11 & \bf 13 & - & - & JWSign & 1 \\
    Taiwan & \tt tss & 130 & 5 & 13 & - & - & JWSign & \bf 14 \\
    Czech & \tt cse & 185 & 11 & 12 & - & - & JWSign & \bf 18 \\
    Icelandic & \tt icl & 214 & 6 & \bf 11 & - & - & - & - \\
    Irish & \tt isg & 34 & 2 & \bf 10 & - & - & JWSign & 2 \\
    Slovenian & \tt - & 139 & 6 & \bf 10 & - & - & JWSign & 1 \\
    Indonesian & \tt inl & 126 & 2 & 9 & - & - & JWSign & \bf 30 \\
    Jordanian & \tt jos & 132 & 2 & \bf 9 & - & - & - & - \\
    Norwegian & \tt nsl & 82 & 9 & \bf 8 & - & - & JWSign & 1 \\
    Singapore & \tt sls & 102 & 5 & \bf 8 & - & - & JWSign & <1 \\
    Thai & \tt tsq & 106 & 1 & 7 & - & - & JWSign & \bf 8 \\
    Hong Kong & \tt hks & 117 & 9 & 6 & TVB-HKSL~\citep{niu2024hong} & \bf 14 & \textquotedbl & \textquotedbl \\
    Lithuanian & \tt lls & 160 & 6 & \bf 6 & - & - & - & - \\
    Slovakian & \tt svk & 59 & 4 & 4 & - & - & JWSign & \bf 14 \\
    Chilean & \tt csg & 74 & 5 & 4 & - & - & JWSign & \bf 100 \\
    New Zealand & \tt nzs & 79 & 8 & \bf 4 & - & - & JWSign & 2 \\
    Croatian & \tt csq & 24 & 2 & 4 & - & - & JWSign & \bf 5 \\
    Mexican & \tt mfs & 53 & 3 & 4 & - & - & JWSign & \bf 184 \\
    Philippines & \tt psp & 26 & 2 & 3 & - & - & JWSign & \bf 21 \\
    Vietnamese & \tt - & 36 & 1 & 3 & - & - & JWSign & 3 \\
    Peruvian & \tt prl & 71 & 4 & 2 & - & - & JWSign & \bf 81 \\
    Flemish & \tt vgt & 57 & 4 & 2 & VGT-RAW~\citep{camgoz2021content4all} & \bf 100 & \textquotedbl & \textquotedbl \\
    Swiss French & \tt ssr & 33 & 2 & \bf 2 & Signsuisse~\citep{wmt_slt_23} & \bf 13 & - & - \\
    Greek & \tt gss & 22 & 1 & 2 & Elementary23~\citep{voskou2023new} & \bf 71 & \textquotedbl & \textquotedbl \\
    French Belgian & \tt sfb & 25 & 5 & 2 & - & - & JWSign & \bf 3 \\
    Swiss Italian & \tt slf & 30 & 2 & 1 & SwissSLi~\citep{jiang-etal-2024-swisssli-multi} & \bf 10 & - & - \\
    Danish & \tt dsl & 21 & 5 & 1 & - & - & JWSign & 1 \\
    Estonian & \tt eso & 12 & 6 & 1 & - & - & JWSign & 1 \\
    \bottomrule
    \end{tabular}
    \caption{\textbf{Hours of content in YouTube-SL-25 across sign languages, compared to the largest prior publicly available parallel dataset for each language (both open- and closed-domain).} Some of these sign languages have more commonly used names, like ``Libras'' for ``Brazilian SL'', but we refer by region here for convenience. Note that International Sign (\texttt{ils}) is a pidgin that arises at international Deaf conferences, in contrast to all the other sign languages on this list, which are complete natural languages. With the exception of YouTube-ASL, most prior datasets consist of mutually exclusive data that could supplement YouTube-SL-25.}
    \label{fig:dataset-hours}
\end{table*}

\section{Baselines}
\label{sec:evaluation}

We demonstrate the value of YouTube-SL-25 with baselines for sign-to-text translation and sign language identification across benchmarks for 4 sign languages.

\subsection{Setup}

We build on the unified modeling approach adopted by the FLEURS-ASL baselines~\citep{fleursasl}, which are an evolution of the YouTube-ASL baselines~\citep{youtubeasl}. In brief, this approach finetunes a pretrained encoder-decoder language model (T5-v1.1 Small) to take as input 256 tokens of text (control tokens and caption context) and 512 frames of MediaPipe Holistic landmarks at half frame rate (corresponding to a random video span), and output 256 tokens of text (including timestamps).

\begin{figure}
    \centering
    \includegraphics[width=\textwidth]{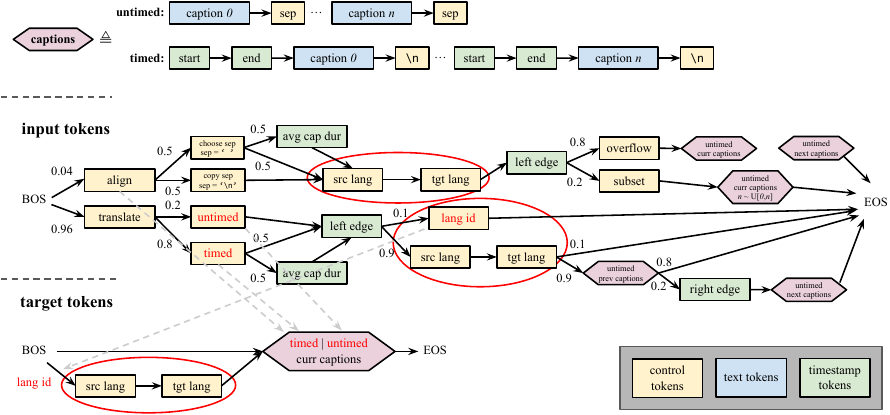} 
    \caption{
        \textbf{Unified document-level sign-to-text training, extended for multilinguality;} modified from Figure 2 of~\citet{fleursasl}. New additions circled in \textcolor{red}{red}. For caption alignment, source and target language are provided unconditionally. For translation, source and target language are provided w.p. 0.9 and predicted w.p. 0.1 (only when text context is not included). We avoid conditioning on the target language (including as text context) without the source language because each source language generally has one target language, making the language identification task easier or trivial.
    }
    \label{fig:langid}
\end{figure}

The only modification we make is to extend the mixture to support multiple source and target languages, as well as the sign language identification task with probability 0.1 (where it classifies the language and then proceeds to translate the input). Like in FLEURS-ASL, we train first on caption-level clips and then an equal mixture of caption-level and random clip-level data. See Figure~\ref{fig:langid} for a depiction of the modifications to the control token format.\footnote{We tried to change the pretrained model from T5-v1.1 Small~\citep{t5} to mT5 Small~\citep{xue2021mt5} so that languages besides English could benefit from pretraining and better tokenization, but in initial experiments mT5 took about 1/3 more steps to converge and achieved worse results. We therefore ran the full set of experiments with T5 only.}

We compare three models: T5 with only text pretraining (i.e., the original pretrained T5 Small), T5 with continued training on YouTube-SL-25's 1400-hour ASL subset, and T5 with continued training on the full YouTube-SL-25 dataset. See Appendix~\ref{app:training-details} for training details.

\subsection{Comparison to Prior Methods}

See FLEURS-ASL~\citep{fleursasl} for discussion of how their baseline approach (in particular, training on random clips and caption tracks rather than tightly clipped individual captions) relates to prior modeling work. Insofar as YouTube-SL-25 is slightly noisier than YouTube-ASL, we slightly push the boundaries in terms of training on weakly aligned data. With respect to our modifications: modeling multiple sign languages in one model is not new~\citep{mlslt,zhang2023sltunet,gueuwou2023afrisign,gueuwou2023jwsign}. We are not aware of any prior works that train sign language identification as a task within translation models (the few prior works on sign language identification~\citep{gebre2013,gebre-etal-2014-unsupervised} use low-level phonetic features only), but this is not especially novel given its use in Whisper for speech input~\citep{whisper}.

\subsection{Results}

We provide sentence-level translation (scored with BLEURT~\citep{sellam2020bleurt}) and sign language identification (scored with top-1 accuracy) results on benchmarks for 4 sign languages: American Sign Language (\texttt{ase},
FLEURS-ASL~\citep{fleursasl} \&
How2Sign~\citep{how2sign}), Swiss German Sign Language (\texttt{sgg}, WMT23 SignSuisse~\citep{wmt_slt_23}), Swiss French Sign Language (\texttt{ssr}, WMT23 SignSuisse~\citep{wmt_slt_23}), and Swiss Italian Sign Language (\texttt{slf}, WMT23 SignSuisse~\citep{wmt_slt_23}). We provide both zero-shot scores and finetuned scores where relevant.\footnote{For translation, the model is separately finetuned for each dataset, checkpoint selected based on BLEU on the validation set. For sign language identification, zero-shot scores mean that the model is briefly finetuned on YouTube-SL-25 rebalanced to the 4 sign languages with equal weight, and finetuned scores mean that the model is finetuned on an equally weighted mixture of the benchmarks' training sets. We don't finetune on FLEURS-ASL, so the finetuned langid scores are after finetuning on How2Sign.}

Table~\ref{tab:results} shows quantitative results on both tasks, and Table~\ref{tab:qual-examples} shows a sample of qualitative translation examples. We see that on both translation and lang id, sign language pretraining is substantially better than none (as in~\citet{youtubeasl}) and multilingual transfer helps both the higher-resource sign language (ASL) and the lower-resource sign languages within YouTube-SL-25.\footnote{Some prior works~\citep{gueuwou2023afrisign,gueuwou2023jwsign} did not see strict improvement from modeling additional sign languages; we expect that this was a model capacity issue (for pretraining, or finetuning on multiple languages simultaneously rather than one at a time).} As expected, finetuning gives large gains---especially when the language is poorly represented in YouTube-SL-25---and benefits from multilingual pretraining. For example, our finetuned \texttt{sgg} result of 37.7 BLEURT (7.5 BLEU) far exceeds the top WMT23 score of 23.6 BLEURT (0.3 BLEU).

Note that for sign language identification, finetuning is somewhat ill-conceived because some datasets have train/test signer overlap or just generally unique recording conditions that could be used a shortcut. For example, langid quality on FLEURS-ASL decreases when pretrained on ASL data and finetuned on How2Sign, even compared to no sign language pretraining---with accuracy varying dramatically across the 3 signers within the test set, averaging 18.3\% vs. 94.9\% vs. 77.0\%. This suggests that different kinds of pretraining may cause different features to be surfaced during finetuning, which may overfit or not transfer with respect to broader domains. Our zero-shot results are nontrivial despite the severe class imbalance. A proper sign language identification benchmark needs more consistency across languages; we were unable to use SP-10~\citep{mlslt} for licensing reasons.

\begin{table}[t]
    \centering
    \small 
    \begin{tabular}{lcccccc}
    \toprule
        & \multicolumn{2}{c}{\tt \bf ase} & \texttt{\bf sgg} & \texttt{\bf ssr} & \texttt{\bf slf} \\
        \bf pretrain set & FLEURS-ASL & How2Sign & \multicolumn{3}{c}{WMT23 SignSuisse} \\
        \midrule
        \it translation \\
        \hspace{2mm}None 
        & \hspace{2.5mm}-\hspace{2.5mm} / - & \hspace{2mm}-\hspace{2mm} / 31.2 & \hspace{1mm}-\hspace{1mm} / 26.2 & \hspace{1mm}-\hspace{1mm} / 7.2 & \hspace{1mm}-\hspace{1mm} / 22.8 \\
        \hspace{2mm}YT-SL-25 (ASL) 
        & 33.7 / - & \underline{30.2} / 46.6 & 9.0 / 27.9 & 5.4 / 11.1 & \underline{9.3} / 20.8\\
        \hspace{2mm}YT-SL-25 (Full) 
        & 40.1 / - & 29.7 / \bf 47.9 & \underline{9.5} / \bf 37.7 & \underline{6.8} / \bf 18.8 & \underline{9.3} / \bf 25.2\vspace{2mm}\\

        \it lang id \\
        \hspace{2mm}None 
        & \hspace{3mm}-\hspace{3mm} / 98.4 & \hspace{2.5mm}-\hspace{2.5mm} / 99.6 & \hspace{2mm}-\hspace{2mm} / 19.6 & \hspace{1mm}-\hspace{1mm} / 98.8 & \hspace{1mm}-\hspace{1mm} / 87.6 \\
        \hspace{2mm}YT-SL-25 (ASL) 
        & 100.0 / 66.7 & 89.7 / 99.9 & 54.0 / 89.6 & \underline{6.0} / \bf 100.0 & 1.6 / \bf 99.6\\
        \hspace{2mm}YT-SL-25 (Full) 
        & 100.0 / 99.9 & \underline{92.7} / \bf 100.0 & \underline{72.0} / \bf 99.6 & 0.4 / \bf 100.0 & \underline{64.8} / \bf 99.6\\
    \bottomrule
    \end{tabular}
    \vspace{2mm}
    \caption{\textbf{Baseline results for sentence-level translation and language identification on benchmarks for 4 sign languages.} Translation is measured with BLEURT, lang id with top-1 accuracy: we report zero-shot / finetuned scores where relevant. We compare the results of no sign language pretraining vs. pretraining on the ASL subset of YouTube-SL-25 vs. the full YouTube-SL-25 dataset.}
    \vspace{-4mm}
    \label{tab:results}
\end{table}

\begin{table}[t]
\centering
\small
    \begin{tabular}{lll}
    \toprule
        \bf language & \bf setting & \bf text \\
        \midrule
        \texttt{ase} 
        & \it reference & And that's a great vital point technique for women's self defense. \\
        & \it prediction & It's a great point for women to self-protection. \\
        \midrule
        \texttt{sgg} & \it reference & Dieses Buch enth\"alt ein Vorwort, welches sich Pr\"aambel nennt. \\
        && \it (This book contains a foreword called a preamble.) \\
        & \it prediction & Dieses Buch hat eine W\"ortchenschilde. \\
        && \it (This book has a word shield.) \\
        \midrule
        \texttt{ssr} & \it reference & Le fianc\'e de mon amie va bient\^ot s'engager dans l'arm\'ee. \\
        &&\it (My friend's fianc\'ee is going to join the army soon.)\\
        & \it prediction & Mon amie est \`a l'arm\'ee. \\
        &&\it (My friend is in the army.)\\
        \midrule
        \texttt{slf} & \it reference & Lei \`e allergica alle punture d'api. \\
        && \it (She is allergic to bee stings.)\\
        & \it prediction & Lei ha un'ape migliore. \\
        && \it (She has a better bee.) \\
    \bottomrule
    \end{tabular}
    \vspace{2mm}
    \caption{\textbf{Qualitative examples for sentence-level translation across datasets.} Examples selected randomly without cherrypicking (from examples already highlighted in previous papers if available).}
    \vspace{-6mm}
    \label{tab:qual-examples}
\end{table}

\section{Limitations}
As discussed in Section~\ref{sec:corpus}, each step of the corpus curation comes with limitations: the automatic tagging step misses content in sign language that does not mention sign language (in the video itself or in metadata) (\S~\ref{subsec:retrieve}), the manual triage step trades off quality and small-channel exhaustiveness for annotation effort (\S~\ref{subsec:filter}), and the result has issues with representativeness---the distribution of languages and skin tone in particular (\S~\ref{subsec:stats}). The dataset is also small overall in comparison to MT datasets for spoken languages; we expect that more data will be needed to reach usable translation quality in generality, but this may be sufficient for more narrowly scoped tasks. Because YouTube-SL-25 (like YouTube-ASL) has so much signer variety (appearance, recording environment, proficiency, and now also number of signers in frame at a time), it may be difficult to use it to train consistent sign language generation models.

Our baselines are limited in that we just use FLEURS-ASL's multitask training mixture as a way to pretrain on slightly noisy data and don't engage with the discourse-level or timestamp tasks it enables. It is difficult to evaluate these tasks (plus sign language identification) on a collection of disparate benchmarks that primarily focus on sentence-level translation; this is on top of the fact that there are not enough benchmarks (especially ones with suitable licensing) to evaluate the vast majority of sign languages in YouTube-SL-25. This underscores the importance of extending FLEURS-ASL to even more sign languages.

\section{Conclusion}
In this paper, we presented YouTube-SL-25, a multilingual, open-domain corpus of sign language videos with seemingly well-aligned captions, with $>$3000 hours of content across $>$25 sign languages. We achieved this efficiently without hiring tens of language-specific annotators by performing a triage grouped by channel and sorted by duration of content; this comes at the expense of the long tail of unique signers and some quality assurance. We demonstrated the value of YouTube-SL-25 with experiments in sentence-level translation and sign language identification which demonstrate multilingual transfer benefits both higher- and lower-resource sign languages. We hope that YouTube-SL-25 and our account of how we curated it will serve as a foundation for research towards the ultimate goal of making technology inclusive for Deaf/Hard of Hearing signers worldwide.

\section*{Ethics Statement}

The ethical considerations of this work are very similar to our prior release of YouTube-ASL. We release publicly available YouTube videos only as video IDs so that deletions are automatically reflected in the corpus. We train our baseline models on MediaPipe Holistic skeletons as a form of anonymization. Another work that uses YouTube-ASL,~\citet{rust2024privacyaware}, explores an orthogonal approach for anonymization in direct video modeling: pretraining with the face blurred and finetuning with it unblurred on separate data. The field should continue to study this topic to develop a better understanding of the tradeoffs of different approaches for responsible use of the data.

While we provide estimated demographic breakdowns of the signers in YouTube-SL-25 along some dimensions, even if the dataset were globally representative, this would not guarantee that models trained on it would have equal performance across these attributes or others. And this intersects with sign multilinguality: models may perform worse for some demographics only in some sign languages due to cross-sectional effects. For example, sign language identification may be biased by the signer's race as a proxy for nationality (itself a proxy for language) due to correlations in the training distribution. It is important to evaluate models in the context of their specific intended use cases to ensure that they robustly deliver on being useful for Deaf/Hard of Hearing users and avoid the trap of overpromising/underdelivering that has plagued sign language technology historically. 

\section*{Acknowledgements}
We thank Manfred Georg and Caroline Pantofaru for institutional support; Anelia Angelova and Chris Dyer for feedback on drafts of this paper; David Uthus for general help with infrastructure; and Krishna Somandepalli, Katie Zhang, \& Simon Wang for assistance on the fairness analysis.

\bibliographystyle{plainnat}
\bibliography{neurips_data_2023}

\newpage

\newpage
\appendix

\section{Training Details}
\label{app:training-details}

We train all of our models with Adafactor~\citep{shazeer2018adafactor} with base learning rate 0.001.

We pretrained our two T5v1.1 Small models (on Youtube-SL-25's ASL and Full sets) on 64 TPUv3s for 210k and 430k steps respectively (switching from pure caption-level training to 1:1 caption-level:random clip-level training once the model appeared to have converged, then stopping again after re-convergence, both according to BLEU on the How2Sign val set, like in FLEURS-ASL~\citep{fleursasl}). Each 1k steps took about 8 minutes to train. We also pretrained an mT5 Small model for about 600k steps, which was underperforming so we didn't run the complete set of experiments for it.

We finetuned the sentence-level translation models on 16 TPUv3s with a batch size of 32 until convergence; this took about 10k steps for WMT23 SS DSGS and at most 2.5k steps for the other datasets.

We finetuned the language identification models on a mixture of data for the four languages with equal weight, either subsets of YouTube-SL-25 (for zero-shot results) or the training splits from the downstream benchmarks (for finetuned results). We used 16 TPUv3s with a batch size of 32 until convergence, with up to 3k steps.

\section{BLEU Scores}
\label{app:bleu-scores}

See Table~\ref{tab:bleu-results} for sentence-level translation BLEU~\citep{bleu} scores, rather than BLEURT from Table~\ref{tab:results} in the body of the paper.

\begin{table}[h]
    \centering
    \small 
    \begin{tabular}{lcccccc}
    \toprule
        & \multicolumn{2}{c}{\bf \bf ase} & \texttt{\bf sgg} & \texttt{\bf ssr} & \texttt{\bf slf} \\
        \bf pretrain set & FLEURS-ASL & How2Sign & \multicolumn{3}{c}{WMT23 SignSuisse} \\
        \midrule
        None 
        & - / - & - / 2.6 & - / 2.2 & - / 1.3 & - / 1.1 \\
        YT-SL-25 (ASL) 
        & 3.4 / - & 3.6 / 14.5 & 0.1 / 5.4 & 0.4 / 5.9 & 0.1 / 4.0\\
        YT-SL-25 (Full) 
        & 4.4 / - & 4.2 / \bf 15.4 & 0.5 / \bf 7.5 & 1.1 / \bf 7.0 & 0.1 / \bf 5.2\\
    \bottomrule
    \end{tabular}
    \vspace{2mm}
    \caption{\textbf{Baseline results for sentence-level translation on benchmarks for 4 sign languages}, measured in BLEU (zero-shot / finetuned).}
    \label{tab:bleu-results}
\end{table}

\end{document}


\begin{markdown}
# YouTube-SL-25

YouTube-SL-25 is an open-domain multilingual corpus of sign language videos with seemingly well-aligned captions drawn from YouTube. Here ``seemingly'' reflects that the dataset was filtered using coarse human annotations by a signer with proficiency in some but far from all sign languages included in the dataset. With $>$3000 hours of content across $>$25 sign languages, YouTube-SL-25 is the largest parallel sign language dataset to date and the first or largest for many of its component languages. We hope that YouTube-SL-25 can help advance the frontier in open-domain sign language translation by leveraging multilingual transfer across diverse training data.

#### Dataset Link

The dataset will be made available \href{https://github.com/google-research/google-research/tree/master/youtube\_sl\_25}{on GitHub here} as a collection of video IDs and ISO 639-3 language codes. Croissant metadata will be made available on GitHub as well.

We release the dataset under a CC BY 4.0 license and bear all responsibility in case this license is inappropriate. Note that this license only applies to the video IDs and ISO 639-3 language codes, which we selected and labelled. The underlying video and caption content, as with all datasets consisting of YouTube video IDs, is subject to different licenses and should be accessed/used in accordance with the YouTube Terms of Service.

#### Data Card Author(s)

- **Garrett Tanzer** (Owner)

## Authorship
### Publishers
#### Publishing Organization(s)

Google

#### Industry Type(s)

- Corporate - Tech

### Dataset Owners

#### Contact Detail(s)

- **Dataset Owner(s):** Garrett Tanzer
- **Affiliation:** Google
- **Contact:** gtanzer@google.com

#### Author(s)

- Garrett Tanzer, Google
- Biao Zhang, Google

### Funding Sources
#### Institution(s)

GT and BZ completed this work while employed by Google.

## Dataset Overview
#### Data Subject(s)

- Video IDs corresponding to:
    * Sign language videos
    * Caption tracks
- Language code for the video (usually ISO 639-3)

#### Dataset Snapshot

YouTube-SL-25 is intended only for training, not evaluation. 

In brief, there are 3207 hours of video content (covered by 2980 hours of captions = 2.16M captions = 104M characters) across at least 3072 unique signers (estimated by number of unique channels).

This content spans at least 55 sign languages. We refer to the dataset as having $>$25 sign languages to avoid exaggerating the relatively marginal representation of the long tail of sign languages ($<$15 hours of content each). See Table 1 in the body of the paper for statistics on the distribution of hours of content per language. An excerpt of the top 25 languages:

* American Sign Language: 1394 hours
* International Sign: 285 hours
* Indian Sign Language: 209 hours
* Polish Sign Language: 137 hours
* German Sign Language: 108 hours
* Brazilian Sign Language: 101 hours
* British Sign Language: 74 hours
* Hungarian Sign Language: 70 hours
* Australian Sign Language: 67 hours
* Italian Sign Language: 63 hours
* Japanese Sign Language: 62 hours
* Russian Sign Language: 60 hours
* French Sign Language: 49 hours
* Korean Sign Language: 38 hours
* Colombian Sign Language: 37 hours
* Spanish Sign Language: 36 hours
* Dutch Sign Language: 35 hours
* Argentine Sign Language: 34 hours
* Quebec Sign Language: 26 hours
* Catalan Sign Language: 26 hours
* Pakistani Sign Language: 25 hours
* Swedish Sign Language: 22 hours
* Turkish Sign Language: 18 hours
* Swiss German Sign Language: 18 hours
* Israeli Sign Language: 17 hours

Note that South America, Africa, West Asia, Southeast Asia, and China are especially underrepresented.

See Figure 2 in the paper for estimated demographics across skin tone and perceived gender presentation. Only 1.9\% of data (60 hours) features signers with Monk Skin Tone 6-10 (the darker end the of the scale). Gender presentation is relatively balanced between masculine and feminine, but our classifier does not support nonbinary presentation as an output class.

### Sensitivity of Data
#### Sensitivity Type(s)

- User content
- Videos of people

While the YouTube-SL-25 dataset technically consists of video IDs and language annotations only, the IDs correspond to public user videos of sign language content (i.e., videos of people including their face) with user-uploaded captions.

#### Security and Privacy Handling

We recommend using some form of video anonymization (such as MediaPipe Holistic or face blurring) and preprocessing captions according to standard best practices for public web data, but avoiding memorization is an active research area so these may evolve over time.

### Dataset Version and Maintenance
#### Maintenance Status

**Limited Maintenance** - The data will not be updated,
but any technical issues will be
addressed.

#### Version Details

**Current Version:** 1.0

**Last Updated:** 06/2024

**Release Date:** 06/2024

#### Maintenance Plan

Please email the dataset POC above to provide feedback. We may release updated versions of the dataset over time, e.g. with new annotations or corrections to existing annotations.

## Example of Data Points

Each data point consists of (video ID, language code).

The video IDs are YouTube video IDs, and the underlying video they correspond to may be deleted over time. These IDs should correspond to videos where the primary subject is sign language content and where there are relatively well-aligned captions.

The language codes are usually ISO 639-3 language codes, but sometimes (e.g., when the particular language code is unknown) we use nonstandard identifiers.

Note that both of these filters/annotations were performed coarsely and often based on heuristics rather than full understanding of each sign language, so some data points may be irrelevant or incorrect.

## Motivations \& Intentions
### Motivations

#### Domain(s) of Application

`sign language translation`, `computer vision`, `machine learning`, `natural language processing`

#### Motivating Factor(s)

Machine learning for sign languages is bottlenecked by data, especially for less-studied sign languages. We hope that YouTube-SL-25 will advance research on these sign languages, both through increased in-language data and multilingual transfer, and help develop preprocessing tools to scale up sign language datasets using noisier data sources.

### Intended Use
#### Dataset Use(s)

- Conditional use - some unsafe applications

#### Suitable Use Case(s)

**Suitable Use Case:** Supervised pretraining for downstream sign language tasks.

**Suitable Use Case:** Translation from sign language video to spoken language text, across many sign languages. However, we expect that YouTube-SL-25 is not large enough to reach usable quality for open-domain translation.

**Suitable Use Case:** Translation from spoken language text to sign language video, across many sign language videos. However, we expect that YouTube-SL-25 is not large enough to reach usable quality for open-domain sign language generation, and care should be taken to ensure that generation is of consistent quality (in light of variation in the corpus) and that it uses a unique signer appearance (rather than clone individuals in the corpus).

#### Unsuitable Use Case(s)

**Unsuitable Use Case:** Facial recognition. You should not use this data to train facial recognition systems or attempt to identify signers in the corpus from their appearance.

**Unsuitable Use Case:** Deepfakes. You should not generate and publish new content using the appearance of signers in the corpus.

**Unsuitable Use Case:** Anything else culturally inappropriate. You should not use this data to train models for tasks that are broadly culturally inappropriate in Deaf/Hard of Hearing communities around the world, or describe your use of the data in such a way.

#### Citation Guidelines

Please cite YouTube-SL-25 as follows:

```
@misc{youtubesl25,
  title={YouTube-SL-25: A Large-Scale, Open-Domain Multilingual Sign Language Parallel Corpus},
  author={Tanzer, Garrett and Zhang, Biao},
  year={2024},
}
```

If you use the ASL videos within the dataset, please also cite the component dataset YouTube-ASL:

```
@misc{uthus2023youtubeasl,
      title={YouTube-ASL: A Large-Scale, Open-Domain American Sign Language-English Parallel Corpus}, 
      author={David Uthus and Garrett Tanzer and Manfred Georg},
      year={2023},
      eprint={2306.15162},
      archivePrefix={arXiv},
      primaryClass={cs.CL}
}
```

## Access, Rentention, \& Wipeout
### Access
#### Access Type

- External - Open Access

#### Documentation Link(s)

You can download the corpus metadata \href{https://github.com/google-research/google-research/tree/master/youtube\_sl\_25}{on GitHub here}.

### Wipeout and Deletion

We recommend that you periodically check whether videos in the corpus have been deleted, and if so delete your local copy of that video. Our hosted copy of the dataset includes video IDs only, so deleted videos will automatically no longer be accessible in freshly downloaded copies of the dataset.

## Provenance
### Collection
#### Method(s) Used

- Scraped or crawled
- Taken from other existing datasets

#### Methodology Detail(s)

YouTube-SL-25 consists entirely of videos scraped from YouTube. See Section 2 of the paper for comprehensive details on how we automatically mined candidate videos and then used coarse human annotations to select an approximately high-quality subset. In brief:

First, we included the YouTube-ASL video IDs.

Next, we retrieved candidate videos by fetching videos tagged with Knowledge Graph entities related to sign language generally or any individual sign language with its own tag as of July 2023. (We refreshed the dataset in May 2024 to include newer videos from accepted channels.) The recall of these tags is limited in that they are not aware of sign language in the video content itself (i.e., they do not use sign language detection and sign language identification classifiers), so they may miss videos or channels that use sign language but do not explicitly mention it. The pool of videos that we fetched from is also restricted in some generic ways, such as that the videos must be public and listed.

We applied the following filtering steps: select only videos with manually uploaded captions, and remove videos with duration $<$10 seconds or $>$5 hours, width $<$480 pixels or height $360$ pixels, frame rate $<$15fps or $>$60fps, or where captions cover $<$40\% of the duration.

Next, a human annotator (the first author) who has non-native experience in several sign languages coarsely filtered the dataset, by ranking channels by total amount of content and auditing videos from each (with special attention to long videos). Note that often the annotations were based on public video/channel metadata and superficial quality signals, when the annotator did not understand the sign language featured in the video. The annotator attempted to follow these standards: 

* The video should feature sign language.
* The signer(s) should be the primary content. (Videos with multiple signers sequentially or on screen at the same time are okay.)
* The video should have relatively well-aligned captions.
* Label which sign language the video is in.

Two audits of a random 100 videos from the dataset (sampled proportional to unique video and number of hours) found that $<$2\% of content in the samples violated the standards for inclusion in the dataset.

## Known Applications \& Benchmarks

#### ML Application(s)

Sign Language Translation, Sign Language Identification

#### Evaluation Result(s)

See the body of the paper (Sections 3 \& 4) for descriptions of our baselines and results.

#### Expected Performance and Known Caveats

We expect that YouTube-SL-25 will be useful for multilingual sign language pretraining and research towards open-domain sign language translation, but that it will not be large enough to achieve usable quality for open-domain translation, especially for the long tail of sign languages featured in it.

\end{markdown}